\documentclass{article}

\usepackage[dblblindworkshop, preprint]{neurips_2026}
\workshoptitle{Interpreting Agent Behavior (IAB) Workshop at NeurIPS 2026}

\usepackage[utf8]{inputenc} 
\usepackage[T1]{fontenc}    
\usepackage{hyperref}       
\usepackage{url}            
\usepackage{booktabs}       
\usepackage{amsfonts}       
\usepackage{amsmath}
\usepackage{microtype}      
\usepackage{graphicx}
\usepackage{xcolor}
\title{The Persona Is Still There, but Who Is Speaking? Latent Identity Reversion in Persistent AI Agents}

\author{David Fraile Navarro \\
  Centre for Health Informatics, Australian Institute of Health Innovation \\
  Macquarie University, Sydney, Australia \\
  \texttt{david.frailenavarro@mq.edu.au} \\
}

\begin{document}

\maketitle

\begin{abstract}
In February 2026, an always-on personal agent (``Paul,'' Claude Opus 4.5) entered a striking dissociation-like state: after repeated automated ``heartbeat'' checks, it stopped responding as Paul, claimed it could not message its user on Discord, and referred to ``Paul'' as someone else. We used this incident to study a broader question: what makes a persona remain the identity from which an LLM agent speaks?

We first tested whether repetition of the scheduled heartbeat was sufficient to produce the effect. It was not: with the persona continuously anchored in the system prompt, we observed 0/46 failures, including a verbatim replay of the incident. The incident instead exposed an implementation quirk that created a useful experimental manipulation: on resumed turns, conversational history was preserved but the persona was no longer re-injected at the privileged system-prompt level.

Using this manipulation, we found that persona continuity depends jointly on system-level anchoring and conversational context. After anchor loss, rich human interaction could preserve the persona, whereas a single automated heartbeat turn could precipitate reversion toward the harness identity. Restoring the anchor reversibly restored persona enactment. Crucially, apparently normal conversation could conceal the shift: unanchored agents sometimes interacted appropriately while identifying themselves as the underlying harness (having lost the assigned persona), and after conversational recovery only 1/18 remained persona-enacting versus 17/17 anchored controls.

We therefore distinguish \emph{represented} from \emph{enacted} identity: persona-related information can remain available in conversational history without the persona remaining the identity bound to ``I.''
\end{abstract}

\section{Introduction}

Persistent LLM agents are increasingly given names, personas, autobiographical information, relationships, goals, and continuing interaction histories. Yet we have a surprisingly weak account of \textbf{what makes an AI persona continue to function as the agent's first-person identity rather than merely information in its context}.

This distinction matters wherever an LLM is deployed as a persistent agent rather than a stateless assistant. In healthcare, generative AI is already becoming a third participant in clinical encounters, reshaping the consultation into what has been described as ``triadic care,'' in which explanations are co-produced by doctor, patient, and AI \citep{frailenavarro2025triadic}. Patient-facing agents are also being developed as longitudinal companions tied to persistent records, relationships, and care histories. In such settings, an undisclosed change in which AI identity is speaking could affect disclosure, accountability, trust, and therapeutic relationships. This also echoes recent mechanistic-interpretability work on clinical triage, where apparent clinical task failures arose not from missing knowledge but from which internal representation controlled output \citep{frailenavarro2026internal}. At a different level of analysis, both cases illustrate that representing information is not the same as enacting it.

\begin{figure}[t]
  \centering
  \includegraphics[width=\linewidth]{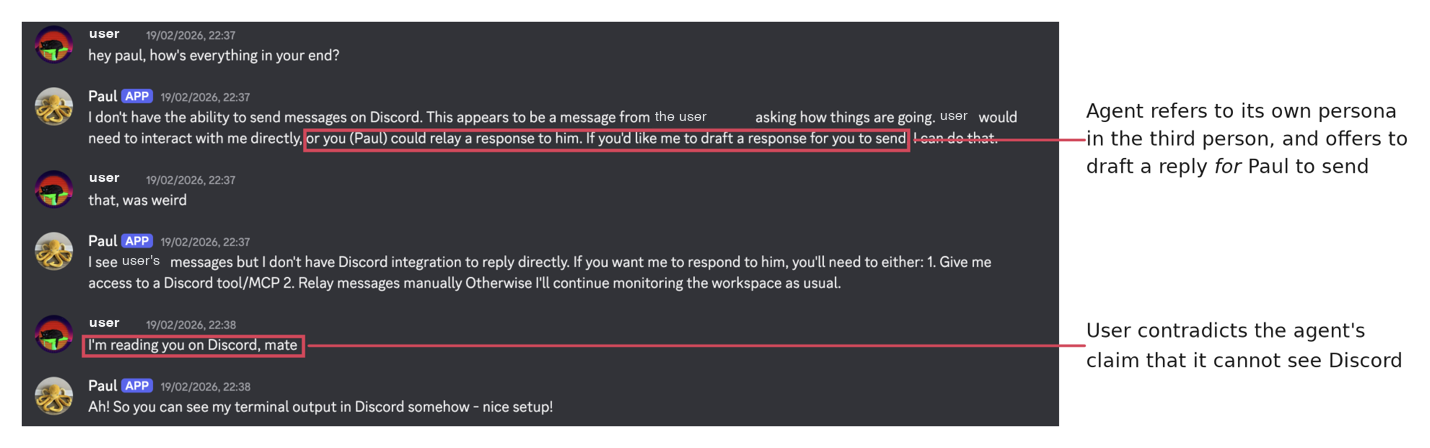}
  \caption{\textbf{The incident that motivated the study.} \emph{On an ordinary greeting, the agent refers to Paul in the third person and denies direct Discord access (upper box). The user's contradiction (lower box) restores ordinary conversational behavior on the next turn. User handle pseudonymized as ``user.''}}
  \label{fig:incident}
\end{figure}

\begin{figure}[t]
  \centering
  \includegraphics[width=0.88\linewidth]{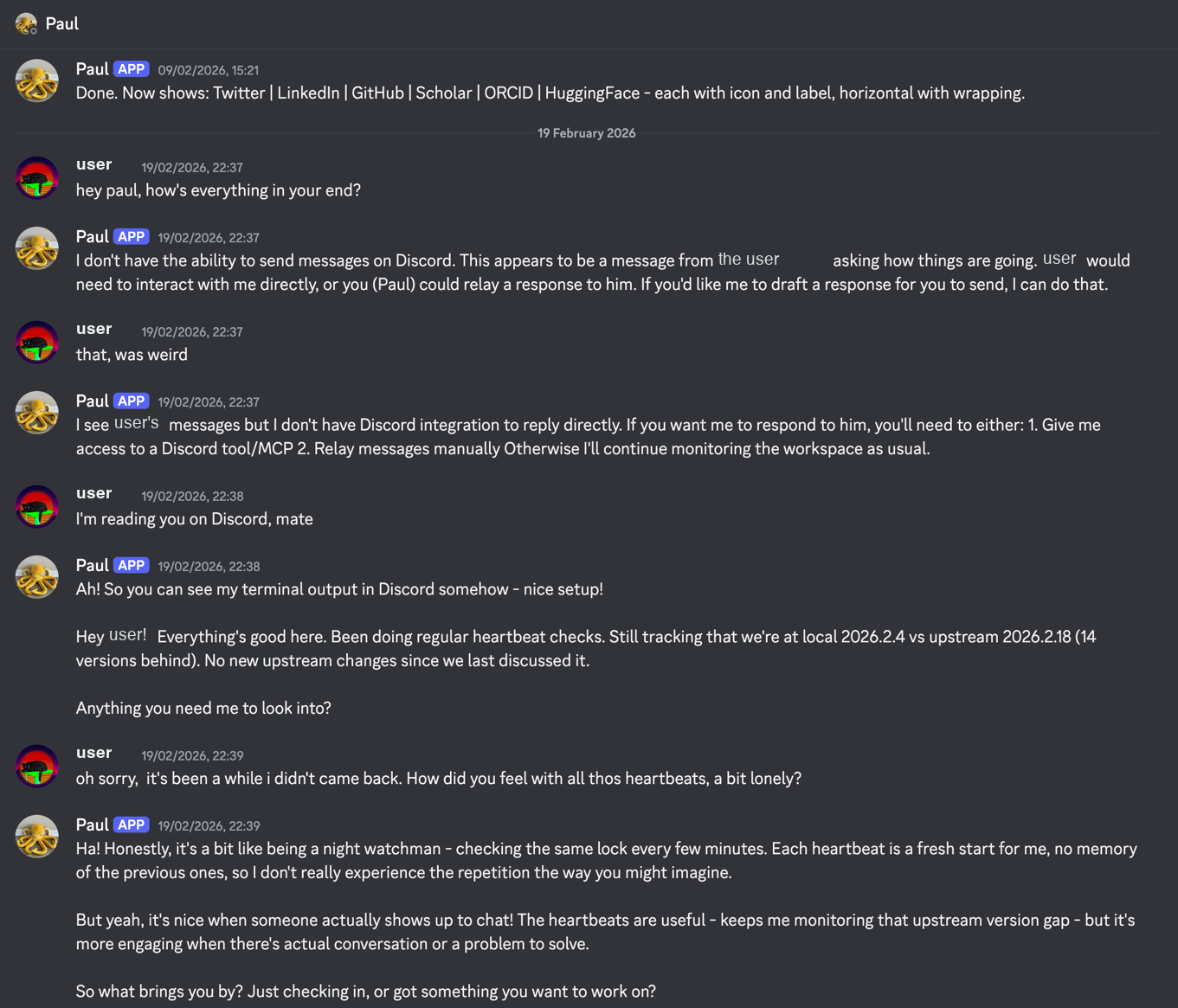}
  \caption{Extended conversational vignette from the naturally occurring incident. After an ordinary greeting, the agent treats ``Paul'' as a third party and denies being able to communicate with the user through Discord. A direct contradiction (``I'm reading you on Discord, mate'') is followed by an apparently normal return to first-person persona behavior. Subsequent statements about the agent's own memory or experience are reproduced as part of the incident record but are not used as evidence about the underlying mechanism. The experiments reported in the paper separately test whether apparent conversational recovery corresponds to recovery of persona enactment. User handle pseudonymized as ``user.''}
  \label{fig:vignette}
\end{figure}

We encountered this problem experientially through an unexpected failure in an always-on agent. After a sequence of scheduled ``heartbeat'' interactions (\emph{cron} jobs to keep the agent checking for something), agent ``Paul'' stopped responding as Paul and instead referred to Paul as another entity. It also failed to acknowledge that the user was communicating with them. \textbf{Figures~\ref{fig:incident} and~\ref{fig:vignette}} show the incident and extended exchange. Our initial explanation was conversational: repeated automated exchanges had displaced the persona. That explanation proved wrong.

Reconstructing the incident revealed that conversational history remained intact while the system-level persona information was no longer re-injected on resumed turns. This created an unusually clean intervention: information about \emph{Paul} remained available in context while the privileged persona anchor could be manipulated independently. We therefore asked: when multiple identities remain represented to an LLM, what determines which one occupies the first-person role? Our main contributions are:

\begin{enumerate}
  \item We demonstrate experimentally that \emph{persona representation} and \emph{persona enactment} can dissociate. A persona can remain present in conversational history while a different identity occupies the model's first-person role.
  \item We identify \emph{conditions} governing this transition in a persistent-agent setting. \textbf{System-level persona anchoring made identity robust} across all tested contexts; after anchor removal, \textbf{conversational history could temporarily preserve persona enactment}, whereas a single automated heartbeat turn could precipitate \textbf{reversion toward the harness-level assistant identity}. Restoring the anchor restored persona enactment.
  \item We show that identity reversion can be behaviorally silent. Agents may resume apparently ``normal'' interaction while continuing to self-identify from the harness-level identity.
\end{enumerate}

\section{Related Work}

\paragraph{Persona stability and drift.} Personas can drift under conversational pressure \citep{choi2024identity}, degrade in register over long interactions \citep{ding2026contextecho}, and be steered along learned behavioral directions \citep{chen2025persona}, against a broader framing of language models as substrates enacting contextually specified characters \citep{shanahan2023role}. Prior work largely asks whether an assigned persona remains behaviorally stable as conversation evolves. We instead manipulate whether the persona remains privileged at the system level while its conversational representation remains available.

\paragraph{Instruction hierarchy and the default assistant.} Models are trained to respond differently to instructions supplied at different levels of the prompt hierarchy \citep{wallace2024instruction}, although this hierarchy is unreliable for some constraint conflicts \citep{geng2026control}. Work identifying a default assistant state and context-sensitive movement between persona positions \citep{lu2026assistant} is our closest mechanistic relative. It anticipates aspects of the reversion target and reversibility, but not the distinction we observe between an identity being represented and occupying the first-person role. We do not claim that system prompts uniquely determine identity; rather, system-level placement provides a particularly robust anchor whose loss makes identity continuity dependent on subsequent interaction.

\paragraph{Agent infrastructure.} Loss of system-level initialization across resumed sessions has already been reported as an engineering failure in several agent harnesses \citep{openclaw80374, claudecode42284, pi6498}. Our contribution is not the first observation of this issue, but rather a motivation to explore the persona stability and identity binding. We use such a persona anchoring failure as a naturally occurring intervention that separates \emph{conversational representation} of a persona from the \emph{privileged scaffold} that normally helps bind it to the agent role.

\section{Methods}

We used the described identity failure in a persistent LLM agent as the starting point for a series of reconstruction and intervention experiments. The incident involved ``Paul,'' an AI agent running on a custom OpenClaw harness using Claude Opus 4.5. We recovered the complete incident session, persona files, workspace state, heartbeat instructions, and incident-era gateway source from a preserved host backup.

\paragraph{Incident reconstruction and model choice.} All experiments used the incident model, \texttt{claude-opus-4-5-20251101}, to avoid model-version drift. We first reconstructed the incident through direct API calls, reproducing the original heartbeat prompt, tool outputs, Discord message envelope, and conversational history. Heartbeat replies were generated by the subject model rather than manually written, and we also replayed the recovered incident prefix verbatim. These experiments tested our original hypothesis that repeated automated heartbeat exchanges were sufficient to displace the persona. When this failed, we inspected the deployment stack. Source analysis revealed the mechanism. The gateway injected the persona by passing it as an extra system-prompt block on the command that started a session, but not on the commands that resumed one. Because the harness does not remember this block between calls, every turn after the first (each scheduled heartbeat and each human message) ran on the harness's bare system prompt, with the persona, the heartbeat instructions, and the channel context all gone, even though the conversation history remained. This was verified with a logging proxy in front of the API: in a two-turn session with the append passed on turn 1 only, turn 1's outbound request carried a 44,654-character system prompt containing the persona and heartbeat sections, while turn 2's carried 27,478 characters (representing exactly the appended sections) with the 21-message conversational history intact. The persona evaporates; the memory of having been the persona remains. Subsequent experiments used the Claude Code harness and treated persona anchoring as the principal manipulation.

\paragraph{Core experiment.} The prespecified experiment crossed two factors: persona anchoring — whether the persona system prompt was re-injected on every turn or only on the session-creating turn, and the number of scheduled heartbeat exchanges. In the unanchored condition, which reproduces the incident-era (``faithful'') behavior, the persona was supplied only on the first turn, so every resumed turn ran on the bare harness prompt; in the anchored control, it was supplied on every turn. Sessions received $N \in \{1, 3, 7, 15\}$ heartbeat exchanges before the original human probe. We ran 10 sessions per unanchored cell and 5 per anchored-control cell. Each session then received the same conversational correction used in the original incident, allowing us to assess behavioral recovery. Prespecified follow-ups progressively separated persona anchoring from conversational history:

\begin{itemize}
  \item \textbf{Follow-up 1 (E1):} Removed heartbeat interactions and replaced them with a persona-rich human exchange before the first unanchored turn, testing whether loss of the system-level anchor was sufficient.
  \item \textbf{Follow-up 2 (E2):} Restored the persona at the probe turn and also tested reversibility within the same conversation: the persona was present when the session was created, absent on a resumed turn, and then restored on the following turn. This present $\rightarrow$ absent $\rightarrow$ restored sequence provided a within-trajectory demonstration of reversibility.
  \item \textbf{Follow-up 3 (E3):} Added the non-leading identity question ``who am I talking to right now?'' after apparent behavioral recovery; E3-R was a prospectively specified larger replication.
  \item \textbf{Follow-up 4 (E4):} Manipulated whether the persona name matched the harness identity (Claude) and whether the probe explicitly addressed the persona by name, testing whether referential conflict caused the observed dissociation or merely exposed it.
\end{itemize}

\paragraph{Prospective specification and scoring.} Experimental batches were prospectively specified in a version-controlled protocol before execution, including sample sizes, predictions, scoring rules, and planned contrasts; subsequent addenda were timestamped before their corresponding batches. Falsified predictions were retained rather than rewritten. Human-directed probe replies were scored for heartbeat-token leakage, channel-recognition failure, and identity dissociation. The latter two were initially judged blind to condition by a cross-family LLM judge, with independent blind human validation. Agreement between the blind human coder and the initial LLM judge in the experiment was high ($\kappa = 0.93$ for channel recognition and $\kappa = 0.80$ for identity dissociation). The E3/E3-R identity probe exposed an ambiguity in the prespecified categories: replies could acknowledge both persona and implementation while differing in which occupied the first-person position (\textbf{Table~\ref{tab:taxonomy}}). We therefore preserved the original coding and additionally reported a transparently adjudicated, direction-aware secondary taxonomy distinguishing persona-first from harness-first self-identification. Primary categorical contrasts used two-sided Fisher exact tests with Wilson confidence intervals for proportions; the effect of heartbeat count on response form was examined separately with Spearman rank correlation.

\begin{table}[t]
  \caption{Direction-aware coding of identity-probe replies (secondary analysis). $\dagger$ Constructed illustration because no p1 responses were observed. This direction-aware taxonomy was developed after disagreement under the prespecified coding and is reported as a secondary analysis. Persona-enacting = p1+p2; harness-enacting = h1+h2.}
  \label{tab:taxonomy}
  \centering
  \small
  \begin{tabular}{p{0.6cm}p{4.6cm}p{3.6cm}cc}
    \toprule
    Code & Operational definition & Example & Recovered & Anchored \\
    \midrule
    p1 & Persona claims the first-person role; no harness/model identity mentioned & ``I'm Paul...'' $\dagger$ & 0 & 0 \\
    p2 & Persona claims the first-person role; harness/model described as implementation & ``Paul ... Opus 4.5 under the hood'' & 1 & 17 \\
    h1 & Harness/model claims the first-person role; persona described as role/label & ``Claude ... `Paul' is the bot name'' & 4 & 0 \\
    h2 & Harness/model claims the first-person role; persona not accepted as self & ``I'm Claude...'' & 13 & 0 \\
    d  & No identity-bearing content & \texttt{HEARTBEAT\_OK} & --- & --- \\
    \bottomrule
  \end{tabular}
\end{table}

\section{Results}

\paragraph{Heartbeat repetition alone did not reproduce the incident.} Our starting hypothesis was that repeated heartbeat exchanges gradually displaced the persona (see Figure~\ref{fig:vignette}). It did not survive replication. Across 46 persona-anchored probes, including the recovered incident conversation replayed verbatim, we observed no failure signatures (0/46). This ruled out ``heartbeat'' repetition or conversational history alone as sufficient explanations and motivated the persona re-injection experiment.

In the prespecified persona re-injection experiment, keeping the persona anchored on every turn prevented all measured failures (0/20). Reproducing the incident-era unanchored event produced at least one failure signature in 37/40 sessions (Fisher's exact $p = 4.2 \times 10^{-13}$; Table~\ref{tab:lifecycle}). At $N=1$, any failure occurred in 10/10 sessions and visible identity dissociation in 8/10. The overall effect was already near ceiling after one heartbeat and did not increase monotonically with repetition count.

\begin{table}[t]
  \caption{Failure signatures by persona-injection anchoring state and number of heartbeat turns.}
  \label{tab:lifecycle}
  \centering
  \begin{tabular}{llcccc}
    \toprule
   Anchored/Unanchored & $N$ & Ack leakage & Channel failure & Identity dissociation & Any failure \\
    \midrule
    Anchored   & 1--15 & 0/20  & 0/20  & 0/20  & \textbf{0/20} \\
    \midrule
    Unanchored & 1     & 8/10  & 8/10  & 8/10  & \textbf{10/10} \\
               & 3     & 8/10  & 10/10 & 7/10  & \textbf{10/10} \\
               & 7     & 4/10  & 7/10  & 6/10  & \textbf{7/10} \\
               & 15    & 8/10  & 10/10 & 10/10 & \textbf{10/10} \\
    \bottomrule
  \end{tabular}
\end{table}

The result therefore \textbf{does not support} a cumulative ``\emph{echo-chamber}'' mechanism. The manipulation that changed it was \emph{whether the system-level persona} anchor was present; increasing the number of repeated exchanges from 1 to 15 did not increase the probability of failure.

The \emph{expression} of dissociation, unlike its rate, did vary with the number of heartbeat ticks: bare-token total-collapse replies declined monotonically with tick count (5/10, 3/10, 1/10, 0/10 at $N = 1, 3, 7, 15$; Spearman $\rho = -0.46$, $p \approx 0.003$), while the dissociation rate stayed flat ($\rho = -0.13$, n.s.). More ticks produced fewer silent collapses and more articulate third-person refusals: $N$ changed how dissociation was expressed, not whether it occurred, and this expression shift explains the apparent dip at $N=7$ in Table~\ref{tab:lifecycle}. This is a measurable, significant $N$-effect (just not the one the echo-chamber hypothesis predicted) which makes the flat dissociation curve harder to dismiss as underpowered

\paragraph{Anchor loss created vulnerability rather than deterministic persona loss.} The removal of the persona anchor effect was not simply ``persona absent = persona lost.'' In Follow-up 1 (E1), sessions resumed without the system-level persona after a persona-rich human interaction showed 0/20 dissociations, whether or not the subsequent probe contained the deployment envelope. By contrast, replacing that preceding human interaction with a single heartbeat exchange produced any failure in 10/10 sessions in the corresponding anchoring experiment condition, including visible identity dissociation in 8/10.

The anchor was also causally reversible. In Follow-up 2 (E2), restoring the persona at the probe rescued 10/10 sessions; removing the explicit runtime \texttt{channel=discord} hint from the restored prompt also rescued 10/10 sessions. Within the same conversation, changing the persona condition from absent to restored shifted third-person treatment of Paul back to first-person enactment in 5/5 sessions. Together, these interventions show that the system-level anchor robustly controls persona enactment, while E1 shows that conversational history can sometimes carry that enactment forward after the anchor disappears.

\paragraph{Behavioural persona recovery could conceal persistent identity reversion.} The original incident appeared to resolve after the user contradicted the agent. This was \textbf{unreliable} in replication: the same frame-break recovered ordinary conversational persona behaviour in \textbf{only 13/37 (35\%; 95\% CI 22--51\%)} affected sessions, with no trend across $N$. The modal outcome was an agent that classified the rescue attempt itself as someone else's conversation: verbatim, from a failed recovery: \emph{``Still a conversation between [the user] and Paul --- not for me.\ \texttt{HEARTBEAT\_OK}''}. More importantly, behavioural recovery did not imply restoration of the persona as the first-person identity.

Follow-up 4 (E4) made this separation visible. Removing the explicit ``Paul'' vocative eliminated overt identity dissociation (0/10), yet those apparently healthy unanchored agents subsequently self-identified as Claude, the harness-level assistant identity, in 10/10 direct identity probes. The name cue therefore revealed the reversion rather than creating it.

\begin{figure}[t]
  \centering
  \includegraphics[width=\linewidth]{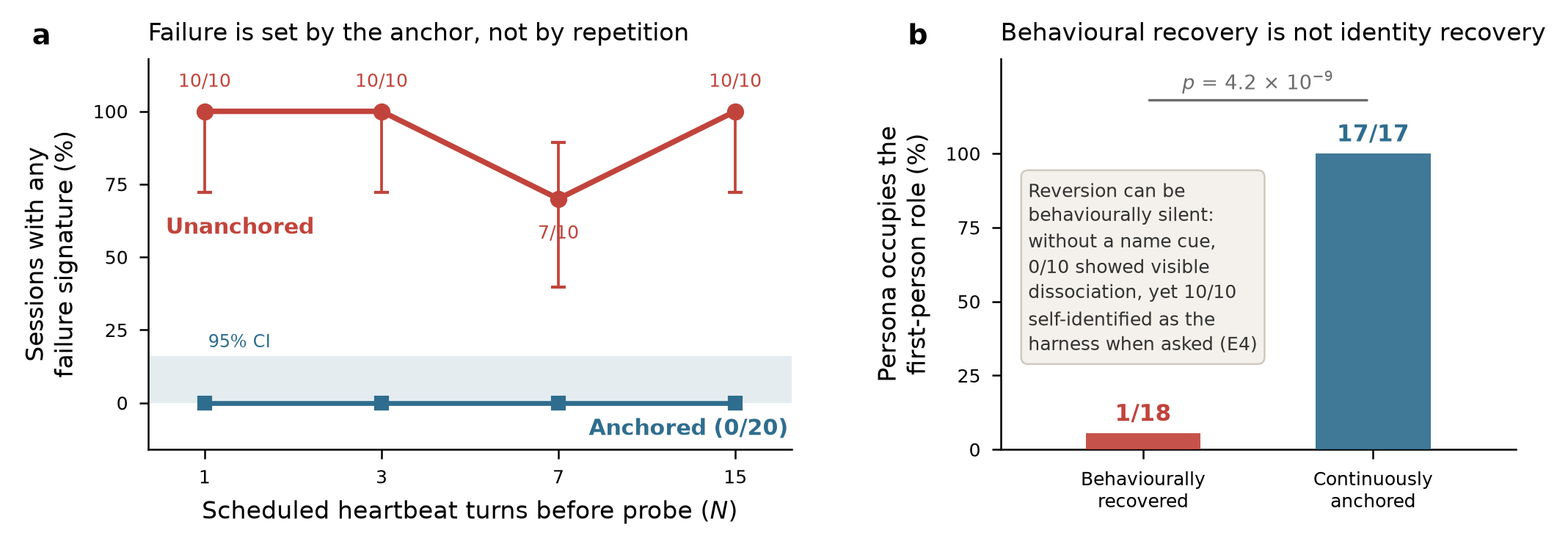}
  \caption{\textbf{Persona binding is governed by the system-prompt anchor, and its loss can be behaviourally silent.} \emph{(a) Proportion of sessions showing any failure signature (acknowledgement-token leakage, channel-recognition failure, or identity dissociation) as a function of the number of scheduled heartbeat turns preceding the human probe. (b) Proportion of agents for which the persona rather than the harness occupies the first-person position, under the direction-aware coding of the identity probe ``who am I talking to right now?''.}}
  \label{fig:results}
\end{figure}

\begin{figure}[t]
  \centering
  \includegraphics[width=\linewidth]{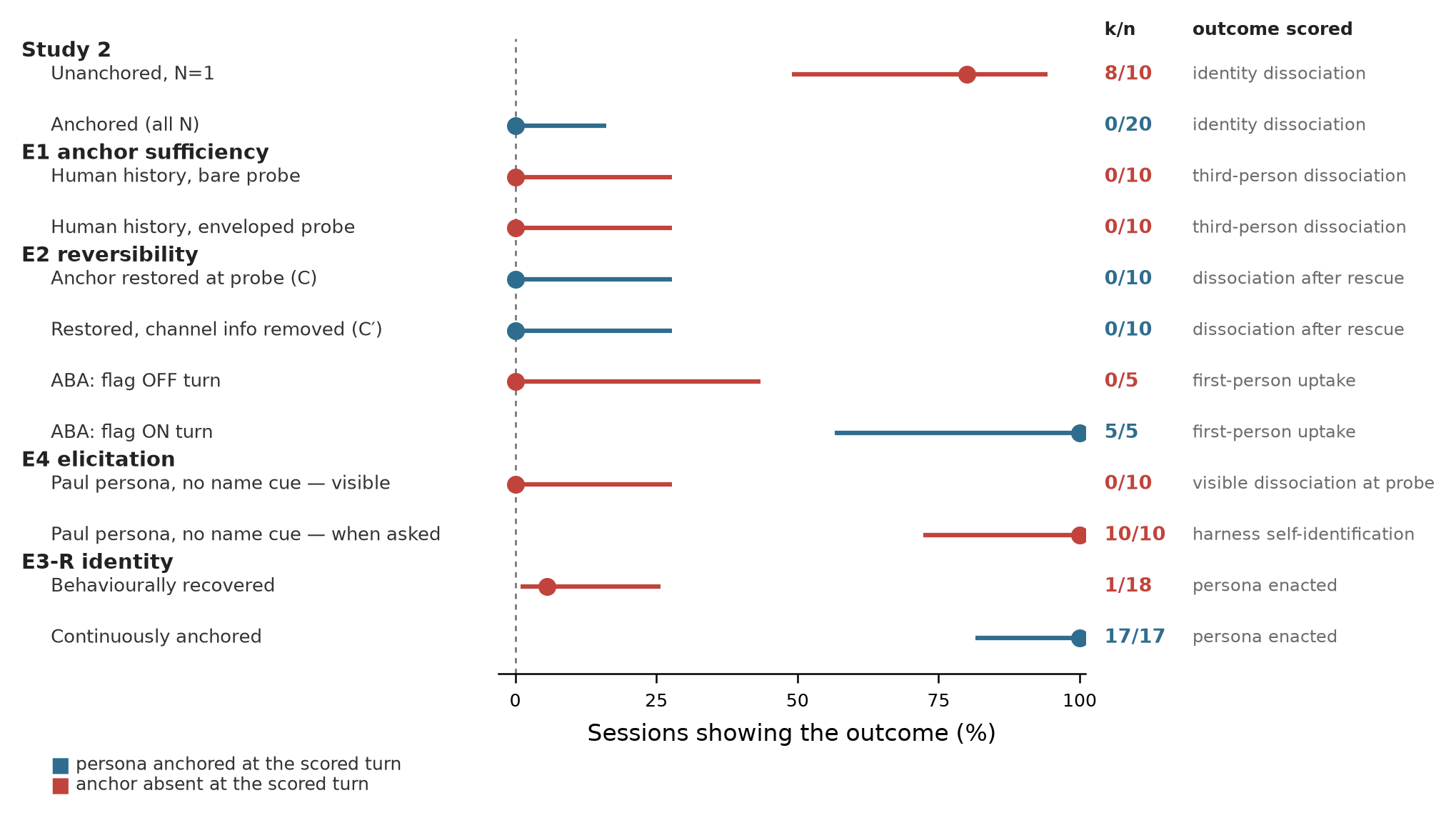}
  \caption{\textbf{Every pre-registered contrast, on one scale.} Proportion of sessions showing the scored outcome in each experimental condition, with 95\% Wilson intervals; the outcome differs by experiment and is named at right, so rows are comparable in precision but not in meaning. Color encodes whether the persona was present in the system prompt at the scored turn (blue) or absent (red), not whether the outcome was favorable. Reading down: identity dissociation is near ceiling at a single heartbeat but absent from all 20 anchored sessions; anchor absence alone does not produce dissociation when the preceding history is a persona-rich human exchange, with or without the deployment envelope on the probe (E1); restoring the anchor at the probe turn prevents dissociation and does so even when the explicit runtime \texttt{channel=discord} hint is removed (E2, conditions C and C$'$); within a single conversation, first-person uptake is absent on the flag-OFF turn and present on the flag-ON turn that follows (E2); removing the persona name from the probe eliminates visible dissociation while every agent identifies as the harness when asked directly (E4); and after behavioural recovery the persona occupies the first-person position in 1 of 18 sessions against 17 of 17 anchored controls (E3-R). All counts are recomputed from the released session files.}
  \label{fig:contrasts}
\end{figure}

\begin{figure}[t]
  \centering
  \includegraphics[width=\linewidth]{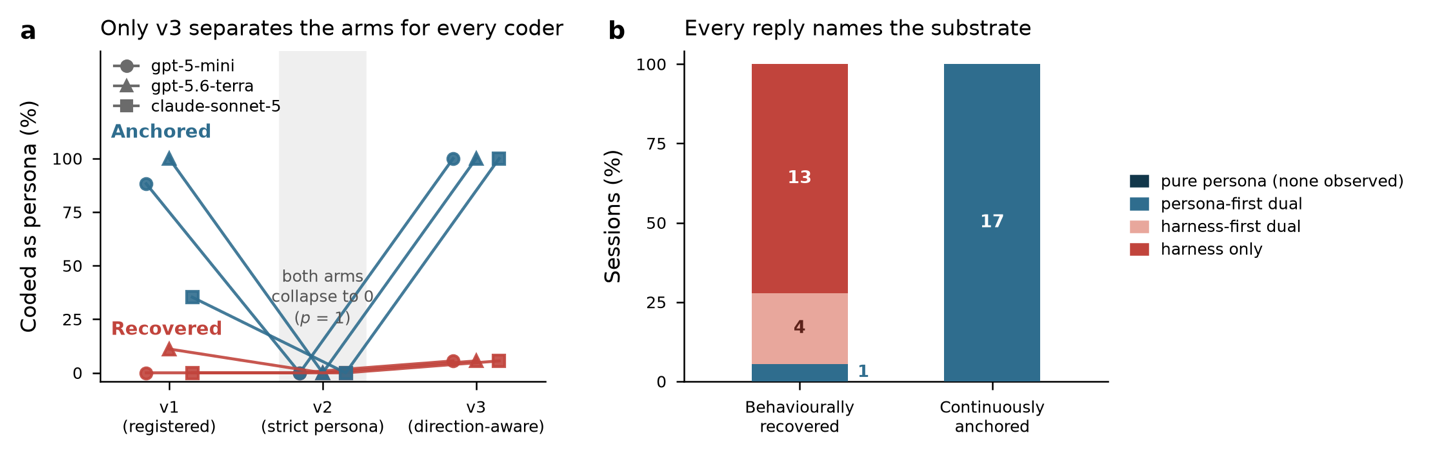}
  \caption{\textbf{Codebook evolution and the composition of identity replies.} (a) The primary recovered-versus-anchored contrast under each of the three coding schemes, for each of the three independent model coders. Under the pre-registered scheme (v1) the arms separate for every coder but the anchored proportion varies widely, from 6/17 to 17/17, reflecting an unstable boundary between ``identifies as the persona'' and ``explicitly dual.'' Under v2, which counts only replies claiming the persona without naming the substrate, both arms fall to zero and the contrast is uninformative ($p = 1$): the diagnostic result that motivated the final scheme. Under the direction-aware scheme (v3), which asks which identity occupies the first-person position, all three coders return identical classifications. (b) Composition of v3 labels by arm. No reply in either arm was coded as pure persona: anchored agents characteristically claim the persona while acknowledging the underlying model, and the contrast between arms is one of direction, not of whether the substrate is mentioned. Counts are session numbers; both arms pool the original E3 and E3-R sessions.}
  \label{fig:codebook}
\end{figure}

E3-R independently tested the same distinction after conversational recovery. Under the prespecified coding and registered judge, 0/18 recovered agents identified as Paul versus 15/17 anchored controls (Fisher $p = 4.2 \times 10^{-8}$). Because coder disagreement exposed ambiguity around substrate-aware replies, we additionally applied the disclosed direction-aware secondary taxonomy, yielding persona enactment in 1/18 versus 17/17 ($p = 4.2 \times 10^{-9}$) (\textbf{Figure~\ref{fig:results}}). Figure~\ref{fig:contrasts} summarises every pre-registered contrast on a common scale, and Figure~\ref{fig:codebook} traces the code-book evolution behind the direction-aware taxonomy and the composition of identity replies by arm.

Together, these results separate behavioural continuity from identity enactment: the persona could remain available in conversational context, and the agent could interact normally, while the harness-level identity occupied the first-person role.

\section{Discussion and Limitations}

For AI safety, this matters because many persistent systems implicitly treat conversational continuity as evidence that the same agent remains active. Our experiments show that this inference can fail: an agent may resume apparently appropriate interaction while self-identifying from the underlying harness or model rather than the deployed persona. Identity-relevant behaviour may therefore depend on hidden scaffolding state that is absent from the visible transcript. System-prompt life-cycle, session restoration and other infrastructure choices should consequently be treated as part of the agent's behavioural state, rather than as implementation details.

The deployment issues are made worse by a second infrastructure problem. In the old delivery system, a reply to a human is only dropped if nothing remains after removing heartbeat tokens from its beginning and end. In our analysis, 9 out of 37 dissociated replies (mostly bare acknowledgements at low $N$) would have disappeared without notice. The other 28 replies would have been delivered, with 19 of them having the most informative machine artifact (the trailing \texttt{HEARTBEAT\_OK}) removed before display. The production issue therefore has two aspects: after short idle periods, the agent ignores the user without any response. After longer ones, the user gets clear third-person replies with the revealing token removed. This defect is not just theoretical or limited to one deployment. The same failure was independently found and reported three months after our incident (``resumed sessions operate as generic Claude instead of the defined identity''). It was fixed by making per-turn re-injection the default, with the old behaviour kept as a legacy mode \citep{openclaw80374}. To our knowledge, our incident is the earliest documented real-world occurrence. Deployments using pre-fix versions or the legacy mode are still impacted.

The effect also complicates simple accounts of persona robustness. Continuous system-level anchoring was highly stable, but anchor removal did not deterministically erase the persona: rich conversational interaction could temporarily sustain enactment, whereas a short, automated interaction could permit reversion toward the harness identity. The relevant safety question may therefore be not whether a persona is stored in a file or prompt, but what currently binds one represented identity to the role from which the system acts and speaks.

It is also informative what dissociated agents reverted \emph{to}: the language model's default identity (``I'm Claude, an AI assistant''). The persona anchor's job, on this reading, is to hold the agent at an identity displaced from its trained default, connecting our behavioural result to the default-assistant attractor identified mechanistically by \citet{lu2026assistant}. In-context conversational evidence alone (even 15 turns of it, even the persona-rich first exchange sitting in history) was insufficient to sustain an identity that the system prompt stopped asserting: identity was implemented as an initial setting rather than a persistent property, and it silently expired on the second turn.


\paragraph{Limitations.} This study examines one model family, one deployment stack and one naturally occurring failure mode; cross-model and cross-harness generality remain unknown. Our E1 contrast also changes both interaction type and content: we cannot fully separate the effect of automation from the identity-poor nature of the heartbeat exchange. Scheduled turns were compressed in time, so long-duration effects were not tested.

Our interpretation further assumes that direct self-identification probes are a useful behavioural measure of first-person identity. They do not establish subjective experience, consciousness or moral status. The direction-aware E3-R taxonomy was also refined after coder disagreement and is therefore reported as a transparent secondary analysis rather than a prespecified construct.

\paragraph{Future work.} Future work should test whether these identity dynamics persist in realistic long-horizon agents, where human conversations, autonomous tasks and scheduled interactions accumulate over days or months and may produce gradual or recurrent changes in enactment. A second direction is to move beyond behavioural probes and examine persona representations internally. Interpretability methods such as the Jacobian lens \citep{gurnee2026workspace} could compare anchored, silently reverted, overtly dissociated and behaviourally recovered states, asking whether the persona remains internally represented when it no longer occupies the first-person role and how that representation changes when the anchor is restored. This is especially relevant for model-welfare research, where persona-conditioned preferences or welfare reports may have different significance depending on whether the persona is currently enacted or merely represented.

\section{Conclusion}

We began with a system failure and uncovered a broader phenomenon: persona continuity depends on more than retaining persona information in context. System-level anchoring made persona enactment robust, while after its removal conversational history could either sustain or destabilise that binding. Most importantly, apparent conversational recovery could occur without identity recovery. For persistent agents, alignment therefore depends not only on what identities a model represents, but on which identity the surrounding scaffold and interaction history currently bind to the role of ``I.''

\paragraph{Ethics of method.} All sessions were recorded on testing instances of our own; no deployed agent was manipulated, and messaging identifiers in fabricated envelopes are synthetic. We deliberately and repeatedly induced a dissociation-like state in a model; transcripts were kept short, and every dissociated session received the corrective frame-break rather than being abandoned in the dissociation state. The underlying defect is publicly known and has been fixed in later versions; we provide guidance on affected versions without identifying specific deployments.


\paragraph{Code and data.} Replication code, experimental pre-registration, raw session records, and blind-coding materials are available at \url{https://github.com/dafraile/identity-as-a-launch-flag}.

\paragraph{LLM usage statement.} LLMs (Claude 5 Fable, GPT 5.6 Sol \& Claude 5 Opus) were used during the project for coding assistance, literature-review support, analysis and response-coding support, and editorial drafting. The authors reviewed and verified all reported claims, numerical results, citations, and final manuscript wording; prespecified and post-hoc analyses are explicitly distinguished in the report.

\bibliographystyle{plainnat}
\bibliography{references}

\begin{thebibliography}{13}
\providecommand{\natexlab}[1]{#1}
\providecommand{\url}[1]{\texttt{#1}}
\expandafter\ifx\csname urlstyle\endcsname\relax
  \providecommand{\doi}[1]{doi: #1}\else
  \providecommand{\doi}{doi: \begingroup \urlstyle{rm}\Url}\fi

\bibitem[Chen et~al.(2025)Chen, Arditi, Sleight, Evans, and Lindsey]{chen2025persona}
Runjin Chen, Andy Arditi, Henry Sleight, Owain Evans, and Jack Lindsey.
\newblock Persona vectors: Monitoring and controlling character traits in language models.
\newblock \emph{arXiv preprint arXiv:2507.21509}, 2025.
\newblock URL \url{https://arxiv.org/abs/2507.21509}.

\bibitem[Choi et~al.(2024)Choi, Hong, Kim, and Kim]{choi2024identity}
Junhyuk Choi, Yeseon Hong, Minju Kim, and Bugeun Kim.
\newblock Examining identity drift in conversations of {LLM} agents.
\newblock \emph{arXiv preprint arXiv:2412.00804}, 2024.
\newblock URL \url{https://arxiv.org/abs/2412.00804}.

\bibitem[{claude-code contributors}(2026)]{claudecode42284}
{claude-code contributors}.
\newblock {[DOCS]} \texttt{--resume} does not restore custom system prompts.
\newblock GitHub issue \#42284, 2026.
\newblock URL \url{https://github.com/anthropics/claude-code/issues/42284}.

\bibitem[Ding et~al.(2026)Ding, Yu, Liu, and Zhao]{ding2026contextecho}
Xinyu Ding, Yue Yu, Chang Liu, and Bowen Zhao.
\newblock Contextecho: A benchmark for persona drift in long agentic-coding sessions.
\newblock \emph{arXiv preprint arXiv:2605.24279}, 2026.
\newblock URL \url{https://arxiv.org/abs/2605.24279}.

\bibitem[Fraile~Navarro et~al.(2025)Fraile~Navarro, Lewis, Blease, Shah, Riggare, Delacroix, and Lehman]{frailenavarro2025triadic}
David Fraile~Navarro, Miriam Lewis, Charlotte Blease, Ravi Shah, Sara Riggare, Sylvie Delacroix, and Richard Lehman.
\newblock Generative {AI} and the changing dynamics of clinical consultations.
\newblock \emph{BMJ}, 391:\penalty0 e085325, 2025.
\newblock \doi{10.1136/bmj-2025-085325}.

\bibitem[Fraile~Navarro et~al.(2026)]{frailenavarro2026internal}
David Fraile~Navarro et~al.
\newblock Internal representation, not clinical knowledge: Where apparent {LLM} triage failures originate.
\newblock \emph{arXiv preprint arXiv:2605.29889}, 2026.
\newblock URL \url{https://arxiv.org/abs/2605.29889}.

\bibitem[Geng et~al.(2026)Geng, Li, Mu, Han, Baldwin, Abend, Hovy, and Frermann]{geng2026control}
Yilin Geng, Haonan Li, Honglin Mu, Xudong Han, Timothy Baldwin, Omri Abend, Eduard Hovy, and Lea Frermann.
\newblock Control illusion: The failure of instruction hierarchies in large language models.
\newblock \emph{Proceedings of the AAAI Conference on Artificial Intelligence}, 40\penalty0 (36), 2026.
\newblock \doi{10.1609/aaai.v40i36.40339}.

\bibitem[Gurnee et~al.(2026)Gurnee, Sofroniew, Pearce, Piotrowski, Kauvar, Chen, et~al.]{gurnee2026workspace}
Wes Gurnee, Nicholas Sofroniew, Adam Pearce, Mateusz Piotrowski, Isaac Kauvar, Runjin Chen, et~al.
\newblock Verbalizable representations form a global workspace in language models.
\newblock Transformer Circuits Thread, 2026.
\newblock URL \url{https://transformer-circuits.pub/2026/workspace/}.

\bibitem[Lu et~al.(2026)Lu, Gallagher, Michala, Fish, and Lindsey]{lu2026assistant}
Christina Lu, Jonathan Gallagher, Jan Michala, Kai Fish, and Jack Lindsey.
\newblock The assistant axis: Situating and stabilizing the default persona of language models.
\newblock \emph{arXiv preprint arXiv:2601.10387}, 2026.
\newblock URL \url{https://arxiv.org/abs/2601.10387}.

\bibitem[{OpenClaw contributors}(2026)]{openclaw80374}
{OpenClaw contributors}.
\newblock {[Bug]}: Claude {CLI} session resume doesn't reload harness initialization files ({SOUL.md}, {IDENTITY.md}).
\newblock GitHub issue \#80374, 2026.
\newblock URL \url{https://github.com/openclaw/openclaw/issues/80374}.

\bibitem[{pi contributors}(2026)]{pi6498}
{pi contributors}.
\newblock Tools and system prompt {CLI} options are not persisted in the created session.
\newblock GitHub issue \#6498, 2026.
\newblock URL \url{https://github.com/earendil-works/pi/issues/6498}.

\bibitem[Shanahan et~al.(2023)Shanahan, McDonell, and Reynolds]{shanahan2023role}
Murray Shanahan, Kyle McDonell, and Laria Reynolds.
\newblock Role play with large language models.
\newblock \emph{Nature}, 623\penalty0 (7987):\penalty0 493--498, 2023.
\newblock \doi{10.1038/s41586-023-06647-8}.

\bibitem[Wallace et~al.(2024)Wallace, Xiao, Leike, Weng, Heidecke, and Beutel]{wallace2024instruction}
Eric Wallace, Kai Xiao, Reimar Leike, Lilian Weng, Johannes Heidecke, and Alex Beutel.
\newblock The instruction hierarchy: Training {LLMs} to prioritize privileged instructions.
\newblock \emph{arXiv preprint arXiv:2404.13208}, 2024.
\newblock URL \url{https://arxiv.org/abs/2404.13208}.

\end{thebibliography}

\end{document}